\documentclass[letterpaper, 10 pt, conference]{ieeeconf}  

\IEEEoverridecommandlockouts                              

\usepackage{graphics} 
\usepackage{epsfig} 
\usepackage{times} 
\usepackage{amsmath} 
\usepackage{amssymb}  
\usepackage{bm}

\usepackage{booktabs} 
\newcommand{\ra}[1]{\renewcommand{\arraystretch}{#1}}
\usepackage{makecell}
\usepackage{multirow}
\usepackage{threeparttable} 

\usepackage{xcolor}

\title{\LARGE \bf
Active Acoustic Sensing for Robot Manipulation
}

\author{Shihan Lu and Heather Culbertson
\thanks{The authors are with the Department of Computer Science, University of Southern California, Los Angeles, CA 90089, USA. {\tt\small \{shihanlu,hculbert\}@usc.edu}}
}

\begin{document}

\maketitle
\thispagestyle{empty}
\pagestyle{empty}

\begin{abstract} 
Perception in robot manipulation has been actively explored with the goal of advancing and integrating vision and touch for global and local feature extraction. However, it is difficult to perceive certain object internal states, and the integration of visual and haptic perception is not compact and is easily biased. 
We propose to address these limitations by developing an active acoustic sensing method for robot manipulation. 
Active acoustic sensing relies on the resonant properties of the object, which are related to its material, shape, internal structure, and contact interactions with the gripper and environment. The sensor consists of a vibration actuator paired with a piezo-electric microphone. The actuator generates a waveform, and the microphone tracks the waveform's propagation and distortion as it travels through the object. 
This paper presents the sensing principles, hardware design, simulation development, and evaluation of physical and simulated sensory data under different conditions as a proof-of-concept. 
This work aims to provide fundamentals on a useful tool for downstream robot manipulation tasks using active acoustic sensing, such as object recognition, grasping point estimation, object pose estimation, and external contact formation detection.
\end{abstract}

\section{Introduction}
Multimodal perception, especially combined visual and tactile sensing, has been trending in robotic manipulation research~\cite{billard2019trends}. One motivation for combining visual and tactile sensing is to complement their respective advantages -- vision can provide better global information while touch can give detailed local information. Visual sensing is non-intrusive, but suffers from occlusion, blur, reflection, delay, and limited accuracy. 
Tactile sensing captures only local shape and surface, but is robust to occlusion and has a higher accuracy for details. Combined visual and tactile sensing can lower the error and delay of state estimation in manipulation. However, this combination is not compact and the perception algorithm design is either complicated or must vary in different experiment setups~\cite{saal2010active}. 
Additionally, it is not possible to complete certain tasks related to the object internal states during the manipulation, 
such as detecting internal structure or cracks, using only vision and touch.

This leads to the question -- can we perceive an object's global states from a localized contact? In daily activities, a person can gauge the water volume in a cup by listening to the resonance due to tapping. Tilting the cup will further change the resonance through the change in water distribution. Through such a local contact between their fingers and the cup, humans can estimate the states such as liquid type, amount, volume, mass, and flow because each state has a distinct resonance. Additionally, tapping position, force, and environment contact will alter the resonance. 

Similar to listening to the resonance after tapping the cup, the method of emitting a vibration waveform and then receiving the waveform after it travels through the object is called active acoustic sensing. Our approach easily integrates this sensing method into existing robot systems by attaching a vibration generator and receiver to separate parts of the robot gripper, without interrupting the regular robotic manipulation routine. It senses the object properties, states, and contact formations that affect the acoustic resonance of the system. Due to the high physical correlation between the received waveform and sensing targets, we can alleviate the dependency on a large amount of data. It also inherits the benefits of active sensing, such as controllable data collection and tight relationship between actions and sensory output.

In this work, we present an active acoustic sensing method for robot manipulation with the following contributions: 
\begin{itemize}
    \item \textbf{Concept} of an active acoustic sensing  method and how it enhances the active perception in robot manipulation tasks with uncertainty
    \item \textbf{Design} of a paired active acoustic sensor on a parallel robotic gripper
    \item \textbf{Simulation} of sensor signals in robot manipulation scenarios based on modal analysis and synthesis
    \item \textbf{Experiment} of the physical and simulated sensor comparison and a proof-of-concept on an object recognition task and a grasping position estimation task
\end{itemize}

\section{Background}
Acoustic sensing, both passive and active, stems from human-computer interaction research for human gesture recognition. Passive acoustic sensing directly uses the produced sounds or structural vibrations to detect a human's gestures or activities~\cite{harrison2008scratch, laput2016viband}. Active acoustic sensing uses an actuator mounted on the hand surface to emit acoustic signals and a receiver on another location of the hand to collect the signals after travelling through the hand~\cite{zhang2018fingerping,kubo2019audiotouch}. Distinct gestures create unique resonant frequencies of the collected signals due to the unique travel paths. In hand-object interactions, with the similar setup combining an emitter and a receiver on hand or object, the contacts between hand and object alter the signal travelling, which can be used to infer gestures~\cite{ono2013touch} or contact forces~\cite{funato2017estimating}. 

In robot manipulation, acoustic sensing is mainly applied to sense the state change of soft pneumatic actuators (SPA), also in both passive and active methods. Z{\"o}ller~\textit{et al.}~\cite{zoller2018acoustic} embedded a microphone into the SPA's air chamber to detect the contact with the environments by induced sounds. They further added a speaker into the chamber to emit a frequency sweep and used the microphone to record the sweep after it travelled through the chamber~\cite{zoller2020active}. In that way, it turned the whole chamber itself into a contact detection sensor by tracking the change of the sweep. A similar setup that combined a speaker and a microphone was proposed by Takaki~\textit{et al.}~\cite{takaki2019acoustic} for measuring the length of an SPA. By replacing the sweep generated by the speaker with the intrinsic noise from the pneumatic system as the sound source, Mikogai~\textit{et al.}~\cite{mikogai2020contact} devised a similar acoustic sensing method for contact point estimation for SPA. This concept has also been extended to the active vibroacoustic sensing, which used the structural vibrations transmitted through the SPA's body, instead of the air sounds~\cite{randika2021estimating}. 

For perception methods in general robotic manipulation that leverage acoustic or vibroacoustic signals, both sounds generated from exploratory actions such as tapping and shaking~\cite{sinapov2009interactive} and vibrations that travel through a tool held by robot~\cite{taunyazov2021extended} were used to decipher the object properties and contact information. Clarke~\textit{et al.}~\cite{pmlr-v87-clarke18a} also estimated the flow and amount of granular materials using the vibrations from the scooping and pouring actions. 
Recently, impact sounds (i.e., the sounds produced by a tool striking an object) were used to infer object properties and contact force profile with an analysis-by-synthesis method~\cite{clarke2022diffimpact}. Those methods showed promise in bridging the key manipulation characteristics, such as contact point, grasping force, and object material or even shape, with related acoustic signals. However, they either required extra exploratory actions besides the regular manipulation routine or were constrained to specific types of objects or tasks. With general grasping tasks, researchers attached a microphone on the gripper to acquire sounds as a cheap alternative to tactile sensing when vision was occluded, but the connection between the sound and contact events was not tight or robust due to the lack of resistance to ambient noises, so it needed certain large amount of data for model learning~\cite{Du-RSS-22}.

\section{Active Acoustic Sensing}
We conceptualize the active acoustic sensing in robot manipulation from the following three perspectives -- sensing principle, sensing scheme, and sensing characterization. We show its unique sensing capabilities and how it differs from and complements other tactile sensing methods. 

\subsection{Sensing Principle}
\begin{figure}
\begin{center}
\includegraphics[width =\columnwidth]{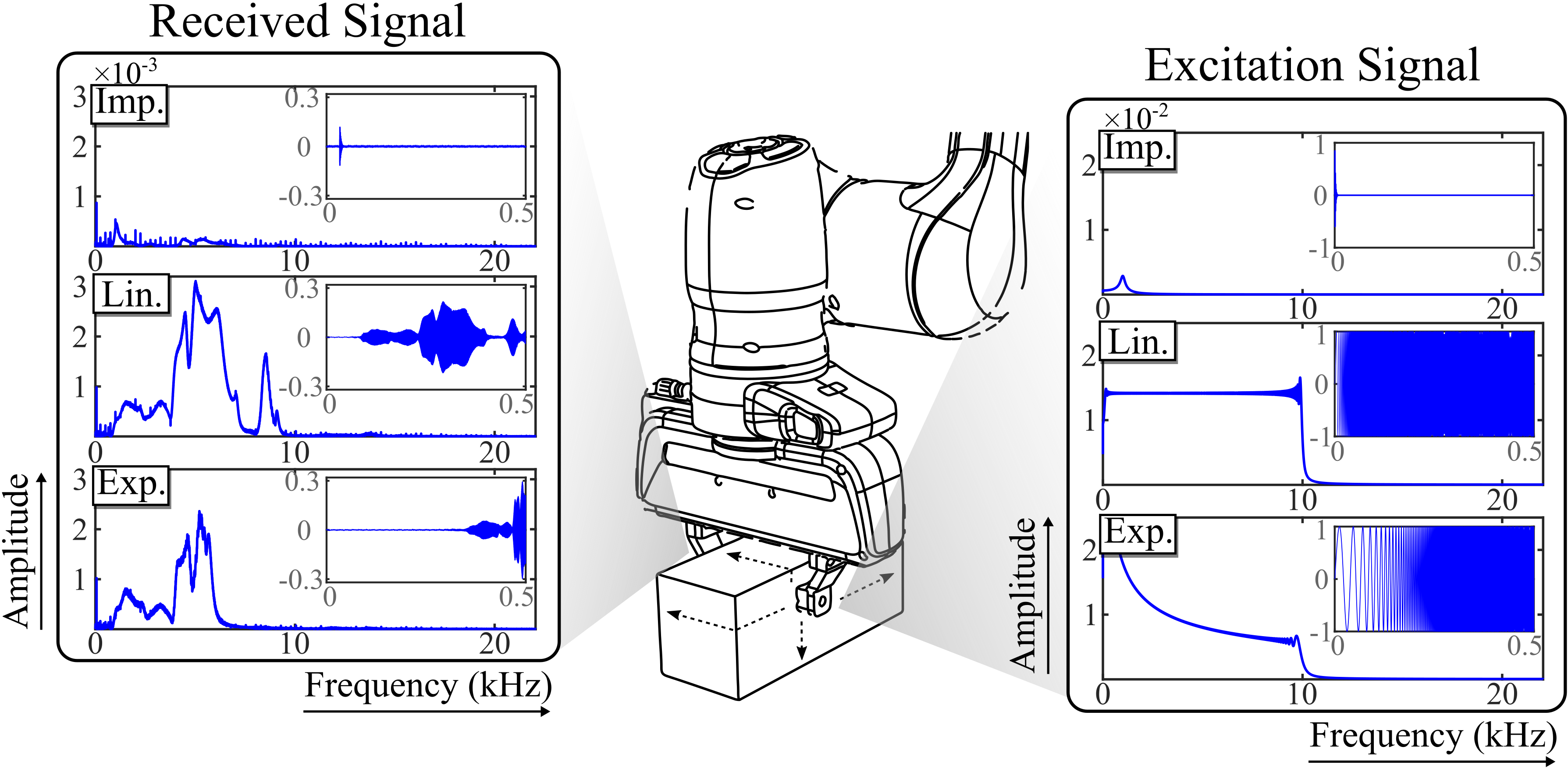}
\caption{Concept of active acoustic sensing for robot manipulation. The actuator on one finger injects a wave into the object and the microphone on another finger records the wave after it travels through the object. Different waves such as impulse, linear sweep, and exponential sweep signals can be applied as excitation signal for different scenarios. Subfigures at top right are raw signals in the time domain (amplitude vs time in seconds).}
\label{concept_active_acoustic}
\end{center}
\end{figure}

By emitting a waveform to the object and capturing the waveform after it travels through the object, active acoustic sensing takes advantage of the phenomenon that the spectral characteristics of the received waveform depend on the acoustic resonant properties of the object under different conditions (Fig.~\ref{concept_active_acoustic}). Different object materials change the medium through which the waveform travels, and different object shapes and grasping points change the path of the waveform. Contact formations with other objects or the environment also affect the waveform propagation by altering the resonant properties such as damping~\cite{wu2021wave}. 

This method is agnostic to light conditions or self-occlusion during the manipulation. It directly connects the local contacts with the related global states such as object shape, material, grasping point, internal state change, and external contact formations. Fundamentally, the proposed active acoustic sensing utilizes the state change of the objects, rather than the state change of the sensor itself, which is called exteroceptive sensing. We admit that it relies on pre-recorded data or simulated data from the object for the state inference, but the high correlation between the sensor data and related object states makes the pre-recording very efficient and the inference model easily generalized.  

Optical-based tactile sensors, such as GelSlim~\cite{donlon2018gelslim}, mainly rely on an object's local surface features to infer the relationship between the object and the gripper during the static contact or pre-lifting stage, which limits its application for objects with homogeneous surfaces. In contrast, active acoustic sensing exploits the object's resonance under excitation for inferring an object's global states regardless of the surface features. 
Additionally, optical-based tactile sensors usually require specifically-designed actions (e.g., shaking, tilting) to implicitly infer the object properties (e.g., liquid type in a container) before performing the actual tasks~\cite{saal2010active,Huang-RSS-22}. Active acoustic sensing directly senses those properties and does not rely on the exploratory behaviors from the robot. This allows it to be seamlessly integrated into the grasping routine, showing a promising direction of augmenting and complementing the optical-based tactile sensing.  

\subsection{Sensing Scheme}
We describe the excitation signal generation and the received signal analysis. Due to the nature of the active sensing, we can control the excitation signal as impulse, linear sweep, exponential sweep signals as input. Different excitation signals offer different frequency bandwidths and resolutions, thus fitting in different tasks for better performance.  

Impulse signals, modelled as a Dirac delta function, are commonly used to analyze the impulse response of an object or a system and often occur as impacts in the real world. Linear sweep signals vary their frequency linearly over time. They sweep through a wide range of frequencies and have low spectral leakage, making them a great choice for determining the object's resonant properties. Exponential sweep signals change frequency exponentially as a function of time. 
Compared to linear sweep signals, they emphasize the low frequencies that may enhance the differentiation between received signals from contacts.

In this work, we emit the impulse signal with duration of 0.01 seconds and use the linear sweep and exponential sweep signals with frequency range from 20 Hz to 10,000 Hz with duration of 0.5 seconds, as shown in Fig.~\ref{concept_active_acoustic}. The frequency range is decided based on the empirical trials with a set of objects, which exhibits notable spectral difference under different conditions. All signals are looped every 0.5 seconds.

We temporally synchronize the signal generation and signal recording threads. To analyze the signals received by the microphone, we first segment the window of 0.5 seconds since all excitation signals are iterated at this rate. For each window, we use Fast Fourier Transform (FFT) to extract the amplitude of the frequency spectrum. Due to the vibration leak from the actuator to microphone through the gripper body and motor noises, we consider only the frequencies from 3 kHz to 10 kHz with a 50 Hz incremental step, resulting in a 140-dimension feature vector. This feature vector can then be fed into tasks such as classification and regression.

\subsection{Sensing Characterization}
\label{sensor_characterization}
\begin{figure}
\begin{center}
\includegraphics[width =\columnwidth]{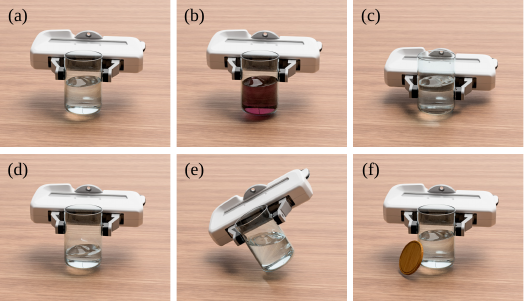}
\caption{
Active acoustic sensors can respond to different object states and contact formations. We show an example of sensing a cup filled with liquid: (a) water cup; (b) different liquid type; (c) different grasping position; (d) different water volume; (e) tilted water cup -- different water distribution in the cup; 
(f) contact with a coaster. All these object states and contact formations have their unique resonant properties.
}
\label{sensing_capability}
\end{center}
\end{figure}

Generally, the core features that can be captured by active acoustic sensing in manipulation are: (1) object material or shape; (2) grasping position; (3) mass and volume of the object in a container; (4) mass (rigid object) and flow (liquid or granular object) distribution of the object in a container; (5) contact formations when interacting with the environment. Using an example of handling a water cup with a parallel gripper, we illustrate the above features in Fig.~\ref{sensing_capability}.

The perception algorithm can be either a hierarchical process, for instance starting with the recognition of the object material and shape, then proceeding to the grasping position and external contact formation estimation, or a centralized model that takes all variables into account at once. In this work, we qualitatively and quantitatively demonstrate features (1), (2), and (5) with a focus on rigid objects.     

\section{Sensor Design}
\begin{figure}
\begin{center}
\includegraphics[width =\columnwidth]{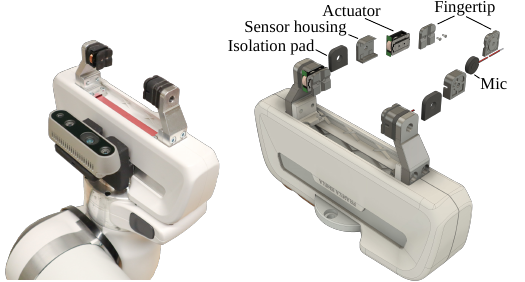}
\caption{Sensor setup on Franka Hand and its explosive view in CAD}
\label{sensor_design}
\end{center}
\end{figure}

This section describes the design of a paired active acoustic sensor on the example of a parallel gripper (Franka Hand). We embed a bone conduction actuator as signal generator and a piezo-electric contact microphone as signal receiver on separate fingers of the gripper. We show the sensor prototype on a Franka Hand and its exploded view in Fig.~\ref{sensor_design}. The compact design can be adapted to other grippers.

\subsection{Actuator and Contact Microphone}
To fit the fingers of most grippers, the selection of actuator and contact microphone for the active acoustic sensor is intended to be compact in size.  
\subsubsection{Bone Conduction Actuator}
On one of the gripper fingers, we use a bone conduction actuator (Dayton Audio BCE-1) to emit waves into the object. Bone conduction actuators convert the electrical signals into vibrations which are transmitted directly to the contact object, instead of amplifying the air around through a membrane such as audio speakers. The actuator covers a wide frequency response up to 19 kHz. With a dimension of 22$\times$14$\times$8 mm$^\text{3}$, the actuator can fit into the finger tip of Franka Hand without sacrificing much grasping width.        

\subsubsection{Piezo-electric Contact Microphone}
On another gripper finger, we attach a piezo-electric microphone (Adafruit), a type of contact microphone, to record the vibration through contact with the object. Contrary to audio microphones, contact microphones are insensitive to sound vibrations in air and only transmit structure vibrations, so they are not affected by ambient noises. The selected microphone, with a radius of 7 mm and thickness of 3 mm, can detect frequencies from 20 Hz to 20 kHz with low noise.

Both actuator and microphone communicate with the computer through an external sound card (Sound Blaster Play! 3; Creative Labs) which samples the signals at 44.1 kHz. The excitation signals are amplified to drive the actuator (2W Class-D Audio Amplifier Board; Sure Electronics). The signal generation and recording are controlled by PyAudio~\cite{pham2006pyaudio}, a Python binding for the PortAudio library. 

\subsection{Sensor Connectors}
The sensors are attached to the gripper fingers using 3D-printed (black resin, Formlabs Form 2) sensor housings with vibration-isolation pads (0.125 in, 60 Duro, Sorbothane). The isolation pads can absorb the vibration transmitted from the actuator through the gripper body and the motor noise from the robot movement. Replaceable fingertips attached to both the microphone and actuator provide an appropriate grasping surface.
The fingertip is screwed to the actuator's front surface through two threaded holes by M1.4 screws and is glued to the microphone using double-sided tape (0.09 mm thickness, AJ Sign). The size of the combined sensors and fingertips is similar to the original rubber tip of Franka Hand. The total cost of a prototype is about 15 US dollars, which is notably more affordable compared to common force or tactile sensors.

\section{Sensor Simulation}
Creating a sensor simulator can greatly accelerate data acquisition, considering the time involved in the physical experiments. Synthesized sensory data are automatically annotated (e.g., grasping point, object pose), so we can easily scale up the size and enhance the diversity of the dataset for learning-based methods for feature extraction in the future. In this work we use existing acoustic analysis and synthesis tools combined with an off-the-shelf physics engine as a fundamental skeleton for our sensor simulator. 

\subsection{Modal-Based Acoustic Analysis and Synthesis}
\label{sec:modal analysis and synthesis}
To simulate the vibration signal received by the microphone at the contact point, we propose to model the signal as the displacement of the object vertex at that location, which depends on the excitation signal, object material and shape, grasping position, and contact formations with the environment. 

\subsubsection{Modal Analysis and Synthesis}
Modal analysis and synthesis is a broadly used method for physics-based sound modeling~\cite{van2001foleyautomatic,o2002synthesizing,ren2010synthesizing} and can be transferred to this task with modifications. Modal analysis determines the \textit{modes}, which are characteristic harmonic vibrations of an object and are considered as independent for small vibrations induced by solid objects. An object has a set of modes, each with its own vibration \textit{frequency}, \textit{decay}, and \textit{mode shape}, which are utilized for the run-time modal synthesis. 

Different modal analysis methods can be applied to compute the modes of an object~\cite{o2002synthesizing,ren2013example}. In this work, we adopt the linear modal analysis method based on object tetrahedral meshes. Assuming small deformations produced by rigid objects, the vibration of an object with $n$ vertices under external forces can be modelled as:  
\begin{equation}
    \mathbf{K} \mathbf{d} + \mathbf{C} \mathbf{\Dot{d}} + \mathbf{M}  \mathbf{\Ddot{d}} = \mathbf{f}
    \label{eq1}
\end{equation}
where $\mathbf{K}$, $\mathbf{C}$, and $\mathbf{M}$ is the stiffness, damping, and mass matrix, respectively, each with size of $\mathbb{R}^{3n\times 3n}$. They are related to object's density, Young's modulus, and Poisson's ratio. $\mathbf{d}\in \mathbb{R}^{3n}$ is the vector of object vertex displacements in 3D, and $\mathbf{f}\in \mathbb{R}^{3n}$ is the vector of external forces in 3D. Using the Rayleigh damping, we can approximate the damping matrix as a combination of the mass and stiffness matrices, such that $\mathbf{C} = \alpha  \mathbf{M} + \beta  \mathbf{K}$, where $\alpha$ and $\beta$ is the mass damping and stiffness damping coefficient, respectively. 

To decouple the system in Eq.~(\ref{eq1}), we solve a generalized eigenvalue problem for the pair $(\mathbf{K}, \mathbf{M})$: 
\begin{equation}
    \mathbf{U}^{\mathrm{T}}  \mathbf{M}  \mathbf{U} = \mathbf{I}, \ \mathbf{U}^{\mathrm{T}} \mathbf{K} \mathbf{U} = \boldsymbol{\Lambda}
    \label{generalized_eig}
\end{equation}
where $\boldsymbol{\Lambda}\in \mathbb{R}^{r\times r}$ is the diagonal matrix with generalized eigenvalues of pair $(\mathbf{K}, \mathbf{M})$ and $\mathbf{U}\in \mathbb{R}^{3n\times r}$ contains the corresponding generalized eigenvectors in column. 
With the Rayleigh damping and substituting $\mathbf{d} = \mathbf{U}  \mathbf{q}$, we can transform Eq.~(\ref{eq1}) to a diagonalized equation:
\begin{equation}
    \boldsymbol{\Lambda}  \mathbf{q} + (\alpha  \mathbf{I} + \beta \boldsymbol{\Lambda})  \mathbf{\Dot{q}} + \mathbf{\Ddot{q}} = \mathbf{g}
    \label{diagonalized eq}
\end{equation}
where $\mathbf{g} = \mathbf{U}^{\mathrm{T}} \mathbf{f}$. We call $\mathbf{U}\in \mathbb{R}^{3n\times r}$ the $r$-mode modal matrix, which represents the conversion between real space and modal space, so $\mathbf{q}\in \mathbb{R}^{r}$ and $\mathbf{g}\in \mathbb{R}^{r}$ can be regarded as the displacement vector $\mathbf{d}$ and the force vector $\mathbf{f}$ in the \textit{modal space}. Similarly, $\mathbf{U}^\mathrm{T}(\alpha \mathbf{M} + \beta \mathbf{K}) \mathbf{U}$ is the modal damping. Based on Eq.~(\ref{diagonalized eq}), the system can be decoupled into $r$ independent ordinary differential equations (ODEs) and each represents a \textit{mode}, 
which can be solved analytically.

The above modal analysis is performed once before the modal synthesis. For the synthesis, when the external forces are applied to the object at different locations, different modes in Eq.~\ref{diagonalized eq} are excited and then summed up to contribute the modal vibration $\mathbf{q}$. Using $\mathbf{d} = \mathbf{U}  \mathbf{q}$, we get the vibration of all vertices caused by the external forces in the real space.    

\subsubsection{Contact-Dependent Damping}
Contrary to the free-body modeling, we also consider the contact-dependent damping, a key factor affecting the vibration output in contact-rich scenarios. Grasping from the gripper and contact with the environment jointly change the object's damping. 

We follow the viscous contact damping model by Zheng~\textit{et al.}~\cite{zheng2011toward} which assumes the existence of viscous dampers at contacts, as depicted in Fig.~\ref{simulation framework}(a). The viscous contact damping is proportional to the contact force extracted from the physics simulation. Given a set of contact points $\mathcal{P}$ with each point $p$, its normal contact direction $\mathbf{n}_{p}\in \mathbb{R}^{3}$, and contact force magnitude $c_{p}$, the viscous contact damping in the modal space can be modelled as:    

\begin{equation}
    \mathbf{G} = \sum_{p\in \mathcal{P}} c_{p} \mathbf{U}_{p}^{\mathrm{T}} (\mu \mathbf{I} + (1-\mu)\mathbf{n}_{p} \mathbf{n}_{p}^{\mathrm{T}}) \mathbf{U}_{p}
\end{equation}
where $\mu$ is the friction coefficient in contact and $\mathbf{U}_{p}\in \mathbb{R}^{3\times r}$ is the sub modal matrix that corresponds to the contact point $p$. 
So, the total damping $\mathbf{C}_{m}$ in the modal space along with the viscous contact damping is:
\begin{equation}
    \mathbf{C}_{m} = \mathbf{U}^\mathrm{T}  (\alpha  \mathbf{M} + \beta  \mathbf{K})  \mathbf{U}   + \gamma \mathbf{G}
    \label{all_damping}
\end{equation}
$\gamma$ is also a material-dependent parameter, like $\alpha$ and $\beta$, for scaling the viscous contact modal damping. Please see \cite{zheng2011toward} for more details on implementation.

\subsection{Simulation Framework}
\begin{figure*}
\begin{center}
\includegraphics[width=\textwidth]{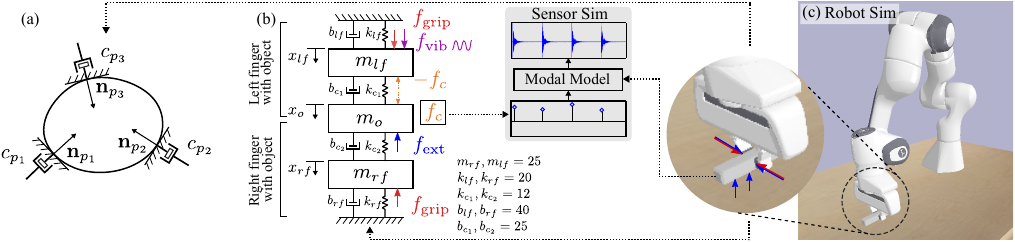}
\caption{Modal-based simulation framework of active acoustic sensing for robot manipulation. (a) Viscous contact damping model in modal analysis; (b) Contact dynamics among object, gripper, and environments for generating collisions;  (c) Robot simulation for extracting high-level contact information.} 
\label{simulation framework}
\end{center}
\end{figure*}

As shown in Fig.~\ref{simulation framework}, our two-pass simulation framework consists of a physics engine for robot simulation and the modal-based sensor simulation introduced above. For the first pass we precompute the modal model from the object meshes. From the robot simulation based on the PyBullet~\cite{coumans2016pybullet} physics engine, we extract the time-series contact dynamic datastream related to the sensor simulation, including contact forces, positions, and directions both from object-gripper pair and object-environment pair. To mimic the collision produced by the actuator, we model the object ($m_o$), gripper's fingertips ($m_{lf}, m_{rf}$), and environment as two-degree-of-freedom second-order mass-spring-damper systems~\cite{huloux2020estimating}, as shown in Fig.~\ref{simulation framework}(b). We compute the collision impulses ($f_c$) between the gripper's left fingertip and the object based on the external force ($f_{\mathrm{ext}}$), grasping force ($f_{\mathrm{grip}}$), and vibration force by the actuator ($f_{\mathrm{vib}}$). All forces are projected to the direction of $f_{\mathrm{grip}}$ of the left fingertip. $k$ and $b$ in Fig.~\ref{simulation framework} are stiffness and damping coefficients in contacts, with subscripts of the corresponding components. 

For the second pass, we synthesize the vibration as the vertex displacement at the contact point between the microphone and the object using the sensor simulation. By taking the datastream from the first pass as input, we update the entire modal damping with the viscous contact damping using all external contacts (Eq.~\ref{all_damping}) and then excite the object using the computed collision impulses (Eq.~\ref{diagonalized eq}). Lastly, we extract the displacement of the vertex that is closest to the contact point as the received signal by the microphone. With this two-pass simulation pipeline, the sensor simulation is invariant to the robot simulation method.   

In physical tests, we observe that there is unavoidable vibration leak from the actuator to the microphone through the gripper body, which will affect the recorded vibration through the object. So, at the idle state of the gripper, we pre-record the leaked vibration by the microphone once for each type of the excitation signal. By aligning the leaked vibration with simulated data via matching their peaks caused by the excitation signal, the leaked vibration is superimposed to the simulated data as the final simulation output.  

\subsection{Limitations}
There are several limitations on the current modal-based sensor simulation framework. First, it is inefficient to synthesize data for high-frequency sweep signals that generate large amount of collisions. Second, since the linear modal analysis relies on the small-scale deformation of rigid bodies, it does not apply to thin-shelled and soft objects. Third, the linear modal analysis also assumes isotropic and homogeneous material across the object, which is not pervasive for daily objects. Fourth, although active acoustic sensing plays an important role in measuring the internal state change, such as the water distribution in a bottle or plush toy's location in a package, such rigid-fluid interaction or rigid-deformable interaction cannot be well handled by this preliminary version. It requires more advanced methods in physics modeling for collision detection, contact computation, and sound propagation in non-rigid objects.

Additionally, annotations of object's acoustic properties are not as common as visual ones. Recent work by Clarke~\textit{et al}.~\cite{clarke2022diffimpact} alleviates the inefficiency of the acoustic property inference by an analysis-by-synthesis method with a differentiable renderer that considers modal responses and force profile. We believe that similar directions such as data-driven methods will greatly enhance the simulation authenticity.  

\section{Experimental Evaluation}
We demonstrate the sensor characterization under different conditions and compare the sensory signals between real and simulated setups. We further evaluate the physical sensor with a Franka Hand on an object recognition task and a grasping position estimation task as proof-of-concept.

\subsection{Experimental Setup}
We tested our physical setup with a planar grasping task using the Franka Emika robot with the Franka Hand equipped with the active acoustic sensor. The grasping force was set at 40 N, which was constant across all conditions. A depth camera (Intel RealSense D435) attached to the hand was used in some tasks for extracting the rough object pose and position. The physical setup was replicated in the simulation, except that the object pose and position were measured directly from the simulator's physics engine instead of the depth camera. For data collection, after a successful initial grasp of the object and before lifting it up, we simultaneously emitted and recorded the signals for 5 seconds using the active acoustic sensor. We only used the first second of each recording for training and testing in our experiments.

\begin{figure}
\begin{center}
\includegraphics[width =\columnwidth]{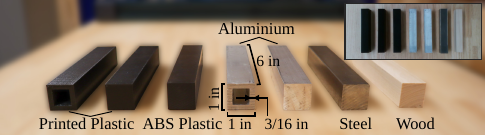}
\caption{Objects used in the experiments. The top right image is the view from the robot's in-hand camera.}
\label{experiment_objects}
\end{center}
\end{figure}

We conducted the experiments using seven geometrical objects with two shapes and five materials, as shown in Fig.~\ref{experiment_objects}. Bar and tube shapes from the same material (aluminium and 3D-printed plastic) have identical visual appearance from the in-hand camera view in planar grasping, so it is almost impossible to recognize them using only vision and it is also very hard to do so with optical-based tactile sensors during the pre-lifting stage. 3D-printed plastic and ABS plastic also share very similar appearance and have minimal visual difference that common cameras hardly distinguish, although their acoustic properties vary significantly due to the micro-structure and material. The material parameters for the simulation are given in Table~\ref{material parameter table}. 3D-printed objects are not simulated due to unknown material properties.

\begin{table}\centering
\ra{1.12}
\caption{Material Parameters for Simulation}
\label{material parameter table}
\begin{tabular}{@{}lcccccc@{}}\toprule
\multirow{2}{*}{\textbf{Material}} & \multirow{2}{*}{\textbf{\thead{Density \\ (kg/m$^\text{3}$)}}} & \multirow{2}{*}{\textbf{\thead{Young's\\ mod. (Pa)}}} & \multirow{2}{*}{\textbf{\thead{Poisson's \\ ratio}}} &  \multicolumn{3}{c}{ \textbf{Damping}} \\ 
\cmidrule{5-7}
           &           &           &       & $\alpha$  & $\beta$  & $\gamma$ \\ \midrule
ABS        &    1100   &      2.6E+9      &   0.36    &    4  &    3E-7    &  4E-2  \\
Aluminium   &    2700   &      6.9E+10     &  0.33     &  0    &   5E-7     &  2E-1 \\
Steel      &    7850   &      2E+11    &   0.29     &   5   &  3E-8  &  3E-1 \\
Wood       &    750    &      1.1E+10  &   0.25    &   60   &  4E-6    &  5E-2   \\ 
\bottomrule
\end{tabular}
\end{table}

\subsection{Validation of Sensor Characterization}
\subsubsection{Setup}
We manually controlled the robot to grasp the object under different conditions. We tested both the physical and simulated system with aluminium bar, aluminium tube, and wood bar using an impulse excitation signal to show the unique sensory signals across the conditions and qualitatively determine how accurate the simulation is at recreating the sensory signals from the physical system. 

\subsubsection{Results}
\begin{figure}
\begin{center}
\includegraphics[width =\columnwidth]{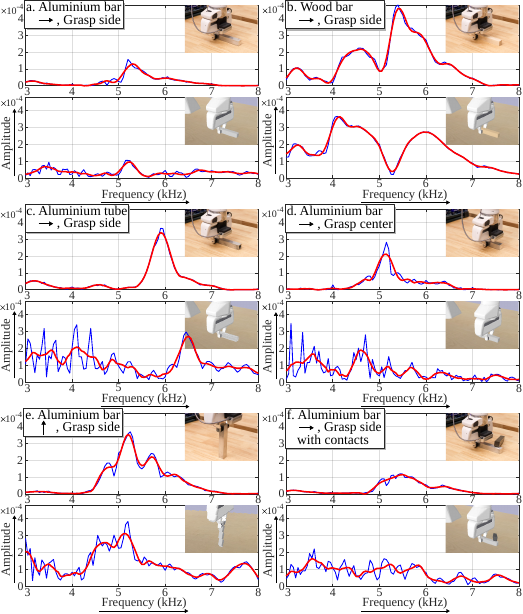}
\caption{Qualitative evaluation of sensor characterizations with both real and simulated data for material, shape, grasping position, and contact formations (contact caused by different poses and contact with other objects). The arrow in the tag means the direction of the object relative to the table. The blue line is the raw spectrum and the red line is the filtered spectrum using Savitzky-Golay filtering. We use 3 kHz to 8 kHz to show the spectral differences.} 
\label{result_sensor_characterization}
\end{center}
\end{figure}

Fig.~\ref{result_sensor_characterization} depicts physical and simulated results for different materials, shapes, grasping positions, and contact formations. 
It is obvious to observe the change in the spectrum of the received signals across the different conditions, which provides the possibility of inferring these states using active acoustic sensing. For (e) in Fig.~\ref{result_sensor_characterization}, although there is no external contact from other objects, the contact with the table caused by different object poses will also affect the sensor data by changing the object damping. This phenomenon can be used to implicitly estimate object poses.  

Most simulated data captured the spectral shape of the real data. The simulated data for aluminium tube failed to mimic the real one since its object shape forced the eigenvalues close to singular thus generated inaccurate results. The condition of aluminium bar contacting a steel bar also produced a large disparity between simulation and reality due to challenges in obtaining accurate contact dynamic values. To obtain realistic simulated data, we need to carefully tune the material parameters $\alpha$, $\beta$, and $\gamma$. Although we manually find those parameters in this work, integrating partial real data into the simulation loop will accelerate this process~\cite{ren2013example}. The quantitative comparison of spectra between real and simulated data depends on the purpose of the comparison, e.g., determining common sources, dominant features, or frequency bin differences. For the generalizability, we omit the quantitative analysis in this paper.  

\subsection{Object Recognition}
We evaluated the sensor's capability to classify different objects through the spectral properties of the received signals. 

\subsubsection{Setup}
\begin{figure}
\begin{center}
\includegraphics[width =\columnwidth]{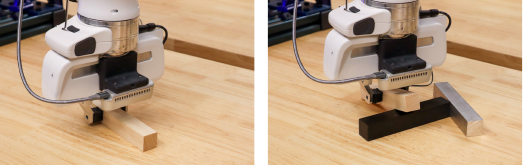}
\caption{Experiment setup for object recognition through contact with the active acoustic sensor. Left: single object without contacts with other objects; Right: multiple objects with external contacts with others.}
\label{experiment_setup_single_multi}
\end{center}
\end{figure}

We evaluated our system with two scenarios: (1) a single object randomly in the scene; (2) multiple objects randomly in the scene with random contacts among the objects (see Fig.~\ref{experiment_setup_single_multi}). The scene with multiple objects in contact has much higher variance and more combinations than the single-object scene, so we evaluate if the classifier trained on single-object scene can be transferred to recognize the object in the multiple-object scene with contacts.   

For the grasping procedure, from the object segmentation from the depth image of the scene, we used the antipodal grasp sampler by~\cite{mahler2017dex} to randomly sample the planar grasping pose with force-closure. We grasped each object 20 times in the single-object scene and 10 times in the multi-object scene for each excitation signal. For training and testing the classifier only in the single-object scene, we conducted the stratified 5-fold cross-validation. For the task of training on the single-object scene but testing on the multi-object scene, we used all the data from the single-object scene for training and all the data from the multi-object scene for testing. 

We predicted the object labels using K-Nearest Neighbors (KNN), Support Vector Machines (SVM), and Multi-Layer Perceptron (MLP) classifiers with \texttt{sklearn} library~\cite{scikit-learn} in Python. For KNN, we selected 3 nearest neighbors with Euclidean distance. For SVM, we used the RBF kernel. For MLP, we set 1 random state with 10$^{\text{-3}}$ learning rate and 1000 iterations, ensuring convergence with our dataset. We used default values for other parameters in the classifiers and retained the same random seed for classification. 

\subsubsection{Results}
\begin{figure}
\begin{center}
\includegraphics[width =\columnwidth]{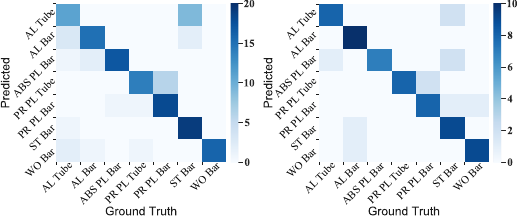}
\caption{Object recognition confusion matrices. Left: single$\rightarrow$single, linear sweep with KNN classifier; Right: single$\rightarrow$multiple, exponential sweep with MLP classifier. We abbreviate the material names to their first two letters.}
\label{experiment_result_1}
\end{center}
\end{figure}

The detailed statistics are shown in Table~\ref{results_for_object_recognition}. For the single$\rightarrow$single condition in the table, the linear sweep reached the overall most robust performance across all three classifiers, with the highest accuracy being 80.0\% by MLP. The exponential sweep achieved the highest accuracy of 82.9\% by MLP in this condition, but had relatively low accuracy by the other two classifiers. 
For the single$\rightarrow$multiple condition, the impulse signal was insufficient to capture the useful features that can be transferred from simple scenarios to complex scenarios, with the highest accuracy slightly over 50\%. 
In contrast, the exponential sweep produced a much better accuracy using all three classifiers, reaching 84.3\% accuracy by MLP. From both conditions, the exponential sweep performed significantly better using an MLP classifier than either KNN or SVM, which illustrates the complexity of the spectral features produced by the exponential sweep. The better performance with exponential sweep also benefits from the greater response precision. From the confusion matrices in Fig.~\ref{experiment_result_1}, we observed that aluminium tube was easily misclassified as steel bar due to the similar metal properties but different shapes. The two shapes of 3D-printed plastic also caused errors in the classification. 

\subsection{Grasping Position Estimation}
We conducted a grasping position estimation task to measure how precisely the sensor responds to spatial differences.  

\subsubsection{Setup}
We horizontally placed the object at a fixed location on the table and controlled the robot to grasp the object at 13 uniformly-spaced positions along half of the main axis with a constant grasping depth. The grasping positions were 5 mm apart and we recorded the data at each position five times. We performed the grasping position estimation as a regression problem using a KNN regressor with three nearest neighbors. We used a 3-fold cross-validation. 

\subsubsection{Results}
\begin{figure}
\begin{center}
\includegraphics[width =\columnwidth]{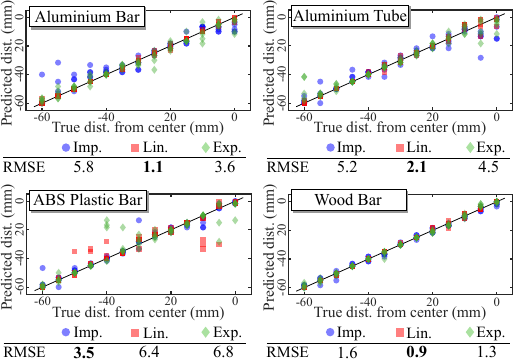}
\caption{Experiment result for grasping position estimation. Best RMSE for each object is in bold.} 
\label{experiment_result_3}
\end{center}
\end{figure}

The results are shown in Fig.~\ref{experiment_result_3}. The black line shows the perfect prediction, so the closer to the black line, the better the prediction is. For all four tested objects, the active acoustic sensing showed good spatial accuracy for the grasping position estimation. 
For three out of four objects, the linear sweep signal outperformed other excitation signals based on the Root-Mean-Square Error (RMSE) between the true distance from the object center and the predicted one.
However, for ABS plastic bar, linear sweep and exponential sweep signals fell behind the impulse signal, which may be caused by the object slip under the intense vibrations due to the slippery plastic surface. 
The test on wood bar achieved the best RMSE of 0.9 mm using the linear sweep.

\begin{table*}\centering
\ra{1.22}
\caption{Experiment Results for Object Recognition}
\label{results_for_object_recognition}
\begin{tabular}{@{}lcccccccccccc@{}}\toprule
\multirow{2}{*}{\textbf{Training$\longrightarrow$Testing}} &  & \multicolumn{3}{c}{\textbf{Impulse}} & & \multicolumn{3}{c}{\textbf{Linear}} & & \multicolumn{3}{c}{\textbf{Exponential}}  \\
\cmidrule{3-5}
\cmidrule{7-9}
\cmidrule{11-13}
&  & \textit{KNN(\%)} & \textit{SVM(\%)} & \textit{MLP(\%)} & & \textit{KNN(\%)} & \textit{SVM(\%)} & \textit{MLP(\%)} & & \textit{KNN(\%)} & \textit{SVM(\%)} & \textit{MLP(\%)}
\\ \midrule
Single$\longrightarrow$Single  &  &    75.7    &   69.3        &   \textbf{81.4}    &      &    78.6       &   77.1    &   \textbf{80.0}  &   &  67.1 &  66.4   &  \textbf{82.9}    \\
Single$\longrightarrow$Multiple &    &  47.1 &      \textbf{55.7}      &    51.4   &       &    67.1  &    60.0    &    \textbf{72.9}    &       &   70.0  &  70.0   &  \textbf{84.3}   \\ 
\bottomrule
\end{tabular}
\end{table*}

\section{Conclusions and Future Work}
We proposed the active acoustic sensing for robot manipulation, with the goal of augmenting and complementing optical-based tactile sensing. Through the local contact with the object, the active acoustic sensing is capable of inferring related global object properties, states, and contact formations. We developed the physical and simulated sensor with a parallel gripper and demonstrated its characterization with experiments. It also produced promising results on the object recognition and grasping position estimation tasks.   

For future work, carefully-designed features will be explored for more complex setups, such as features related to the phase of the received signal, although the naive features based on Fast Fourier Transform in this paper performed well in our experiments. We also expect to refine the simulation framework with differential renderers or data-driven approaches to better mimic the real-world data. This will allow the creation of a large-scale active acoustic sensing dataset for various objects with diverse robot manipulation tasks, for instance grasping, inserting, and pouring. It will facilitate learning-based methods for feature extraction of acoustic signals in manipulation. Based on the promising experiment results, we plan to integrate this sensing method into the robot control loop to tackle the scenarios when the precise visual feedback is not available. In conclusion, we aspire that this endeavor can provide a new perspective on challenging perception problems in robot manipulation.








\bibliographystyle{IEEEtran}
\bibliography{references}

\end{document}